# Grasping versus Knitting: a Geometric Perspective
## Comment on
## Hand synergies: Integration of robotics and neuroscience for understanding the control of biological and artificial hands, by M. Santello et al.

Grasping an object is a matter of first moving a prehensile organ at some position in the world, and then managing the contact relationship between the prehensile organ and the object. Once the contact relationship has been established and made stable, the object is part of the body and it can move in the world. As any action, the action of grasping is ontologically anchored in the physical space while the correlative movement originates in the space of the body. Robots—as any living system—access the physical space only indirectly through sensors and motors. Sensors and motors constitute the space of the body where homeostasis takes place. Physical space and both sensor space and motor space constitute a triangulation, which is the locus of the action embodiment, i.e. the locus of operations allowing the fundamental inversion between world-centered and body-centered frames. Referring to these three fundamental spaces, geometry appears as the best abstraction to capture the nature of action-driven movements. Indeed, a particular geometry is captured by a particular group of transformations of the points of a space such that every point or every direction in space can be transformed by an element of the group to every other point or direction within the group. Quoting mathematician Poincaré, the issue is not find the *truest* geometry but the *most practical* one to account for the complexity of the world[1]. Geometry is then the language fostering the dialog between neurophysiology and engineering about natural and artificial movement science and technology. Evolution has found amazing solutions that allow organisms to rapidly and efficiently manage the relationship between their body and the world[2]. It is then natural that roboticists consider taking inspiration of these natural solutions, while contributing to better understand their origin.

The recent European project The Hand Embodied is a remarkable application of this multidisciplinary research paradigm[3].

The human hand is certainly the most sophisticated organ evolution has provided to allow a living system to act on the world. The hand is recognized as a fundamental component of intelligence. Its richness comes from its extraordinary capacity to perform a large range of dexterous manipulation tasks ranging from hammering to knitting. Hammering required maintaining a stable grasp between the handle and the moving arm. How to configure all the degrees of freedom of the hand around the handle, and what configurations obey the better the physical constraints of hammering (e.g. non sliding and force resistant contacts)? They are challenging questions even for a simple hammering task. Knitting is a task much more complicated that hammering. Knitting requires mobile dexterity. Finger movements have to be coordinated in order to steer the thread along the needles while tuning its tension. With respect to hammering, knitting adds another level of challenge. Complexity arises both from the dimension of the hand control space and from the dimension of the task as defined in the physical space.

---

[1] H. Poincaré, L'espace et la géométrie, Revue de métaphysique et de morale, 1895, vol. III, p. 631-646.

[2] A. Berthoz, Simplexity: Simplifying Principles for a Complex World, Yale Univ. Press, 2012.

[3] M. Santello et al. Hand synergies: Integration of robotics and neuroscience for understanding the control of biological and artificial hands, Physics of Life Reviews, 2016.

Grasping an object goes back to establish and maintain a fixed relation between an arbitrary object frame and an arbitrary hand frame. Grasping task and its physical constraints is then described in the physical space by the space of hand placements, i.e. a space of dimension 6. Considering the many possible postures of the hand (i.e. the many placement of the fingers around the object), the question of grasping is to select the ones that fitful the constraints. The link to be settled is between a six dimensional space and the hand high dimensional configuration space. Many hand postures might be admissible and many movements reaching an admissible posture might be feasible. The hand is said to be redundant with respect to grasping task. Redundancy requires methods for posture and movement selection. Posture and movement spaces are highly dimensional spaces. Their dimensions give a measure of the computational complexity of the task. All current researches in life science as well as in engineering explore how living and artificial systems face such a complexity. Both understanding the computation foundations of actions performed by a human hand, and devising a human-like hand impose overcoming the famous curse of dimensionality. This dual perspective is supported by the notion of synergy. Synergies are a way to reduce the dimension of spaces to be explored. At first glance, we may consider two types of synergies. Postural synergies reduce the dimension of the system configuration space. They are the consequence of physical links as cams, or they are derived from holonomic differential links that impose integrable constraints between the velocities of some body parts. Motor synergies tend to reduce the scope of all possible movements by orchestrating all elementary motor controls from a basis whose dimension is less than the dimension of entire motor control space. What makes the outstanding quality of the Pisa/IIT SoftHand is exactly the clever combination of postural and motor synergies.

It is natural that roboticists consider taking inspiration of evolution. It is not mandatory. Evolution did not invent the wheel, and aircrafts do not flap wings, at least at this moment! To face the world complexity, engineering has developed its own concepts without reference to any living system. For instance, the concept of minimalism refers to the methodological approach to design the least complex solutions for a given class of tasks, by, e.g., using the minimal number of actuators or control variables, or the simplest set of sensors. Applied to grasping tasks, this perspective gave rise to clever mechanisms such as a gripper based on the jamming of granular material[4] or a general-purpose hand made of only three rigid fingers[5]. Such general principles percolate with the concept of synergies or with the so-called morphological computation[6].

We have seen that actions are defined in the physics space while their very origin takes place in the control space. The relationship between "action in the real world" and "movement generation" in the motor control space is defined in terms of geometry. It particularly derives from differential geometry, linear algebra and optimality principles[7,8]. Optimal control is based on well-established mathematical machinery ranging from the analytical approaches initiated by Pontryagin[9] to the recent developments in numerical analysis[10]. It allows action-driven movement generation. Such mathematical machinery can operate also in a reverse perspective to elucidate the laws of natural movements and to reveal motor synergies. For instance, inverse optimal control is a way to model the human motion in terms of controlled systems: given an underlying hypothesis of a system, as well as a set of observed natural actions recorded from an experimental protocol performed on several participants, the question is to find the cost function the system is optimizing. From a mathematical point view, the inverse problem is much more challenging than the direct one. There are few (recent) results in this direction of research. They include numerical analysis[11], statistical analysis[12] and also part of the active area of machine learning[13,14].

---

[4] E. Brown, N. Rodenberg, J. Amend, A. Mozeika, E. Steltz, M. Zakin, H. Lipson and H. Jaeger, Universal robotic gripper based on the jamming of granular material, Proceedings of the National Academy of Sciences, Vol. 107, No. 44, 2010.

[5] M. T. Mason, A. Rodriguez, S. Srinivasa, A. S. Vazquez, Autonomous Manipulation with a General-Purpose Simple Hand, The International Journal of Robotics Research (IJRR), Vol. 31, No. 5, April, 2012, pp. 688-703.

[6] R. Pfeifer, J. Bongard, How the body shapes the way we think: a new view of intelligence, The MIT Press, 2007.

As final comment, let us consider the following issue. Imagine a humanoid robot whose arms are equipped with two Pisa/IIT SoftHands as prehensile organs. Is such a robot capable of knitting? The answer is no. The bottleneck comes from synergies. We have seen that the advantage of synergies is to allow space dimensionality reduction, and then complexity reduction. In return, not all possible movements can be generated. Knitting would require facing against the synergies used for grasping. Unraveling the current synergy orchestration would be the first step of a solution. The second step would be to learn new synergies dedicated to knitting. And that is out of the scope of current know-how from robotics and life science. Now, another question is: what is the interest for a robot to learn knitting whereas Joseph Marie Jacquard invented the mechanical loom in 1801? As many human know-how since the beginning of the technology, [great-]grandmother knitting know-how is disappearing.

---

[7] E. Todorov, Optimality principles in sensorimotor control, Nature Neuroscience, Vol. 7, 2004.

[8] J.P. Laumond, N. Mansard, J.B. Lasserre, Optimization as Motion Selection Principle in Robot Action. Communications of the ACM, Vol. 58, N. 5, 2015.

[9] L. Pontryagin et al. The Mathematical Theory of Optimal Processes, Interscience, Vol. 4, 1962.

[10] J.F. Bonnans, J.C. Gilbert, C. Lemaréchal, Numerical Optimization: Theoretical and Practical Aspects, Springer, 2006.

[11] M. Diehl, K. Mombaur (Eds), Fast Motions in Biomechanics and Robotics, LNCIS 340, Springer, 2006.

[12] T. Inamura, Y. Nakamura, I. Toshima, Embodied Symbol Emergence based on Mimesis Theory, Int. J. of Robotics Research, Vol. 23 (4), 2004.

[13] T. Mitchell, Machine Learning, McGraw Hill, 1997.

[14] J. Kober, J. Bagnell, J. Peters, Reinforcement learning in robotics: A survey. The Intern. J. Robotics Research, Vol. 32(11), 2013.